\documentclass[letterpaper, 10 pt, conference,dvipsnames]{ieeeconf}  
\IEEEoverridecommandlockouts
\usepackage[utf8]{inputenc}
\usepackage[T1]{fontenc}
\usepackage[export]{adjustbox} 
\usepackage{diagbox}
\usepackage{colortbl} 
\usepackage{color}
\usepackage{xcolor}
\usepackage{commath}
\usepackage{tabularx}
\usepackage{diagbox} 
\usepackage{booktabs}
\usepackage{cancel}
\usepackage{bm}

\usepackage{graphicx} 
\usepackage{amsmath}
\usepackage{amssymb}  
\usepackage{cancel}
\usepackage{verbatim}
\usepackage{algorithm}
\usepackage{algpseudocode}
\usepackage{mathtools}
\usepackage{float}
\usepackage{tikz}
\usetikzlibrary{shapes.geometric}
\usetikzlibrary{decorations.pathmorphing}
\usetikzlibrary{arrows.meta, positioning, shapes}
\usepackage{comment}
\usepackage{subcaption}
\usepackage{hyperref}
\usepackage{siunitx}
\usepackage{soul}
\usepackage[percent]{overpic}
\usepackage{multirow}
\usepackage{booktabs}

\usetikzlibrary{automata,positioning}
\tikzstyle{node}=[align=center]
\let\labelindent\relax
\usepackage[inline]{enumitem}

\title{\LARGE \bf
Connectivity-Aware Representations for Constrained Motion Planning via Multi-Scale Contrastive Learning}

\author{Suhyun Jeon$^{1}$,  Yumin Lim$^{1}$, Woo-Jeong Baek$^{2}$, Hyeonseo Kim$^{3}$, Suhan Park$^{4,\dagger}$, and Jaeheung Park$^{1,5,\dagger}$% <-this % stops a space
\thanks{*This work was supported by the National Research Foundation of Korea (NRF) and the Institute of Information \& Communications Technology Planning \& Evaluation (IITP), both funded by the Korea government (MSIT) (Nos. RS-2024-00461583, RS-2024-00459435).}%
\thanks{$^{1}$Suhyun Jeon, Yumin Lim and Jaeheung Park are with the Department of Intelligence and Information, Graduate School of Convergence Science and Technology, Seoul National University, Republic of Korea. {\tt\small [suhyun0606, ckrgksakdmac, park73]@snu.ac.kr}}%
\thanks{$^{2}$Woo-Jeong Baek is with the Artificial Intelligence Institute (AIIS), Seoul National University, Republic of Korea. {\tt\small wjbaek@snu.ac.kr}}%
\thanks{$^{3}$Hyeonseo Kim is with the Department of Mechanical Engineering, Seoul National University, Republic of Korea. {\tt\small smtamh@snu.ac.kr}}%
\thanks{$^{4}$Suhan Park is with the Department of Robotics, Kwangwoon University, Republic of Korea. {\tt\small park94@kw.ac.kr}}%
\thanks{$^{5}$Jaeheung Park is also with ASRI, AIIS, Seoul National University and the Advanced Institutes of Convergence Technology, Republic of Korea.}%
\thanks{$\dagger$ Equally advising}%
}

\makeatletter
\def\namedlabel#1#2{\begingroup
    #2%
    \def\@currentlabel{#2}%
    \phantomsection\label{#1}\endgroup
}
\makeatother

\begin{document}
\maketitle
\begin{abstract}
    The objective of constrained motion planning is to connect start and goal configurations while satisfying task-specific constraints. Motion planning becomes inefficient or infeasible when the configurations lie in disconnected regions, known as essentially mutually disconnected (EMD) components. Constraints further restrict feasible space to a lower-dimensional submanifold, while redundancy introduces additional complexity because a single end-effector pose admits infinitely many inverse kinematic solutions that may form discrete self-motion manifolds. This paper addresses these challenges by learning a connectivity-aware representation for selecting start and goal configurations prior to planning. Joint configurations are embedded into a latent space through multi-scale manifold learning across neighborhood ranges from local to global, and clustering generates pseudo-labels that supervise a contrastive learning framework. The proposed framework provides a connectivity-aware measure that biases the selection of start and goal configurations in connected regions, avoiding EMDs and yielding higher success rates with reduced planning time. Experiments on various manipulation tasks showed that our method achieves 1.9 times higher success rates and reduces the planning time by a factor of 0.43 compared to baselines.
\end{abstract}

\section{Introduction} \label{sec:intro}

\begin{figure}[t]
    \centering
    \hfill
    \includegraphics[width=0.48\textwidth,trim={0 0 0 0},clip]{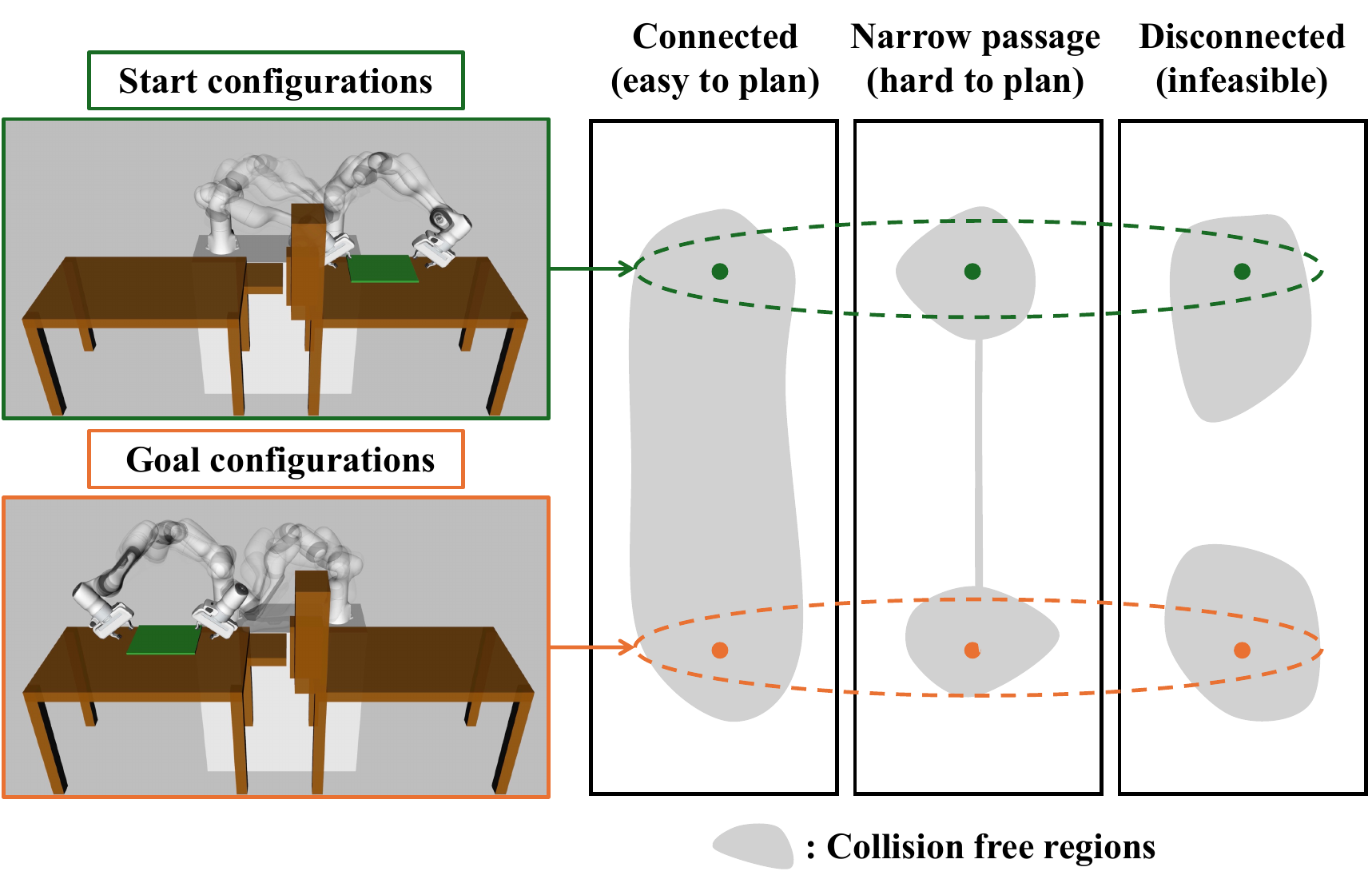}
    \caption{Illustration of how different configuration selections in constrained motion planning result in varying levels of planning difficulty. (Left) Multiple inverse kinematic solutions for the start and goal pose. (Right) Connected configurations are easy to plan, configurations in narrow passages are difficult to plan, and disconnected configurations are infeasible.}
    \label{fig:main}
\end{figure}

Constrained motion planning refers to the problem of computing a feasible path between a start and a goal configuration, where the entire path lies on the constraint manifold \cite{stilman2007task}. Such constraints commonly arise in robotic manipulation, for example, in bimanual or contact-rich tasks that require closed-chain kinematics, contact maintenance, and other physical or geometric conditions \cite{cohn2024constrained, liu2022robot, pettinger2024efficient, krebs2022bimanual}. These conditions restrict feasible configurations to a low-dimensional manifold embedded in the high-dimensional joint space. While this causes difficulties for conventional planning techniques like sampling-based planners in \cite{orthey2023sampling}, addressing these challenges is essential for enabling robots to perform complex manipulation tasks in real-world environments.

A central difficulty in constrained motion planning is that the exact structure of the constraint-induced manifold is unknown. Traditional approaches have focused on extending sampling-based planners to handle such constraints \cite{berenson2009manipulation, jaillet2012path, kim2016tangent, kingston2018sampling} through manifold projection, geodesic distances, or local planners designed to respect constraints. More recently, learning-based approaches have been proposed to approximate constrained manifolds directly, allowing motion planning to be guided by latent representations \cite{qureshi2020neural, park2024constrained}. Although these efforts have advanced constrained motion planning, they have focused mainly on improving sampling efficiency and designing local planners. 

In redundant manipulators, a single end-effector pose admits infinitely many inverse kinematic solutions that may form continuous self-motion manifolds or fragment into discrete components \cite{cohn2024constrained, burdick1989inverse}. Given a start pose and a goal pose, each admits multiple inverse kinematics solutions, yielding many start–goal configuration pairs whose connectivity is unknown. As illustrated in Fig.~\ref{fig:main}, different choices of start and goal configurations lead to different planning performances: (i) connected configuration pairs can be solved quickly, (ii) configuration pairs separated by planner-hard environments, such as narrow passages, require significantly longer search time, and (iii) truly disconnected configuration pairs lead to planning failure. Following Xian et al.~\cite{xian2017closed}, cases (ii) and (iii) are instances where the configurations lie in essentially mutually disconnected (EMD) components. EMD components are defined as regions of the configuration space that are either truly disconnected or practically unreachable, such that no feasible path can be found within a reasonable planning budget.

This work introduces a framework that learns connectivity-aware representations of the configuration space through multi-scale contrastive learning. The objective is to construct a feature-space where shorter distances between points correspond to a higher probability of connectivity. To achieve this objective, the proposed method proceeds in two stages. First, manifold learning is applied at multiple neighborhood scales to obtain embeddings, and each embedding is clustered to generate pseudo-labels. Second, these pseudo-labels supervise a neural network via multi-scale contrastive learning, where pseudo-labels from different scales are jointly incorporated so that the feature-space reflects connectivity information across multiple levels of granularity. The resulting representation is used in the pre-planning stage of constrained motion planning: multiple start–goal inverse kinematic candidates are embedded, and the closest pair in the learned feature-space is selected. This process leads to a higher likelihood of generating feasible plans, improving success rates and reducing planning times.

The main contributions of this paper are as follows:
\begin{itemize}
   \item This paper proposes a simple and efficient method for generating multi-scale pseudo-labels for constrained motion planning, utilizing manifold learning across neighborhood scales and clustering to capture the structure of the configuration space.
    
   \item The proposed method learns connectivity-aware representations that guide the pre-planning stage to select start–goal configurations more likely to find feasible paths efficiently.
    
    \item Comprehensive experiments on diverse manipulation tasks validate the effectiveness of the proposed method, demonstrating higher success rates and reduced planning times compared to baseline methods.
\end{itemize}

The remainder of this paper is organized as follows. Section \ref{sec:related} reviews related works. Section \ref{sec:problem} formulates the problem statement. Section \ref{sec:method} presents the proposed method in detail. Section \ref{sec:experiments} reports the experimental setup and comparative results. Finally, Section \ref{sec:conclusion} concludes the paper.

\section{Related Works} \label{sec:related}

\subsection{Sampling-Based Constrained Motion Planning}
Sampling-based planners such as RRT~\cite{lavalle1998rapidly, kuffner2000rrt} and PRM~\cite{kavraki1998analysis} have been widely extended to constrained settings. Typical strategies include enforcing loop-closure constraints through projection or biasing sampling toward configurations that satisfy task-specific constraints~\cite{berenson2009manipulation, jaillet2012path, kim2016tangent, kingston2018sampling, stilman2010global, berenson2011task}. These approaches improve the exploration of low-dimensional restricted configuration spaces. However, a fundamental limitation remains that sampling-based planners cannot certify that no feasible solution exists~\cite{orthey2023sampling}.

This limitation becomes more severe in closed-chain problems ~\cite{yakey2002randomized, cortes2002random}, where loop-closure constraints can fragment the feasible space into disconnected components ~\cite{burdick1989inverse, xian2017closed}. Although strategies such as regrasping or contact reconfiguration may restore feasibility ~\cite{xian2017closed, lertkultanon2018certified}, they increase planning complexity and are not always applicable. Alternative approaches attempt to mitigate these issues by diversifying the set of candidate goal states, for example, through multi-goal pose selection~\cite{hernandez2019lazy} or iterative addition of inverse kinematic solutions~\cite{jang2022motion}. Nevertheless, these methods do not explicitly consider the connectivity of the constraint manifold. In particular, feasibility depends on whether the start and goal states lie in the same connected region. This gap highlights the need for connectivity-aware representations.

\subsection{Manifold Approximation and Representation Learning}

Data-driven approaches have recently emerged to approximate constrained manifolds directly, enabling planners to generalize feasible regions without explicit reliance on geometric models~\cite{park2024constrained, qureshi2020neural, qureshi2021constrained, acar2021approximating, sutanto2021learning}. These methods improve sampling efficiency and feasibility prediction, but they primarily capture local validity and do not explicitly address whether configurations are connected or not. These limitations motivate the need for representation learning methods that are connectivity-aware and capable of preserving global topological properties of the configuration space.

In parallel, representation learning has gained increasing traction in robotics~\cite{cho2024corn}, inspired by its success in vision and natural language processing. Contrastive learning, in particular, has proven highly effective at learning structured embeddings from images, point clouds, and videos~\cite{chen2020simple, khosla2020supervised, li2021contrastive}. However, its application to robot configuration spaces remains limited because explicit connectivity labels are unavailable. As a result, most prior efforts in robotics have been restricted to perception-oriented tasks, and no existing work has leveraged contrastive learning to explicitly encode connectivity for motion planning. To address this gap, we generate pseudo-labels via clustering to supervise contrastive learning over configuration embeddings.

\begin{figure*}[t]
    \centering
    \includegraphics[width=0.98\textwidth,trim={0 0 0 0},clip]{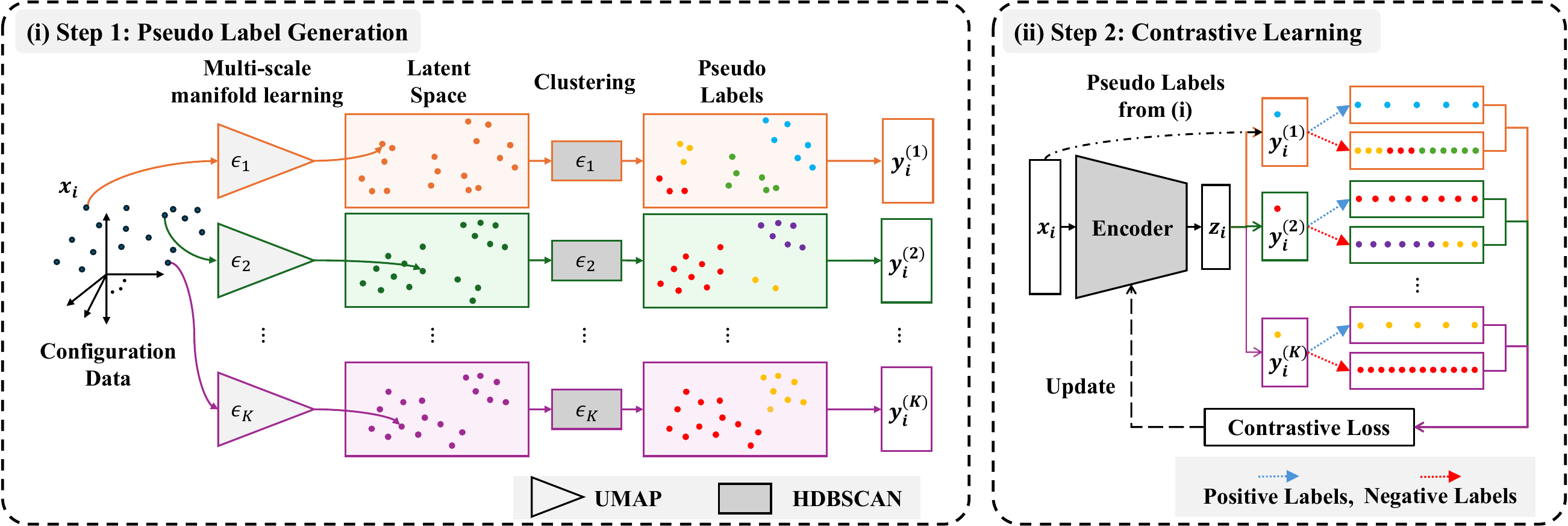}
    \caption{Overview of the proposed method. The pipeline consists of two stages: (i) multi-scale pseudo-label generation via manifold learning and clustering at different neighborhood scales, and (ii) multi-scale contrastive learning guided by these labels, where configurations sharing the same pseudo-label are pulled closer in the feature-space and those with different labels are pushed apart, resulting in a connectivity-aware representation.}
    \vspace{-0.2cm}
    \label{fig:method}
\end{figure*}

\section{Constrained Motion Planning: Closed-Chain Constraints} \label{sec:problem}

Multi-arm manipulation tasks with closed-chain constraints can be expressed in terms of the relative pose between end-effectors.  
Let $T_i(q) \in SE(3)$ denote the end-effector pose of the manipulator $i$ at configuration $q \in \mathcal{C}_{\mathrm{robot}}$.  
The relative transform between manipulators $i$ and $j$ is
\[
T_{ij}(q) = T_i(q)^{-1} T_j(q).
\]
  
Let $\mathcal{C}_{\text{free}} \subset \mathbb{R}^n$ denote the collision-free configuration space of the multi-arm system.  
The closed-chain constraint can be expressed as
\[
T_{ij}(q) = T_i(q)^{-1} T_j(q) = T_{ij}^d, \quad \forall (i,j) \in \mathcal{G},
\]
where $\mathcal{G}$ is the set of manipulator pairs grasping the object and $T_{ij}^d$ is the desired relative pose.  
The corresponding feasible manifold is
\[
\mathcal{M} = \{ q \in \mathcal{C}_{\text{free}} \mid T_{ij}(q) = T_{ij}^d, \forall (i,j) \in \mathcal{G} \}.
\]

Given object-scale start and goal poses $x_{\mathrm{start}}, x_{\mathrm{goal}} \in SE(3)$, the inverse kinematics solution sets on the manifold are
\[
Q_{\mathrm{start}} = \{ q \in \mathcal{M} \mid T_{\mathrm{obj}}(q) = x_{\mathrm{start}} \}
\]
\[
Q_{\mathrm{goal}} = \{ q \in \mathcal{M} \mid T_{\mathrm{obj}}(q) = x_{\mathrm{goal}} \}.
\]

The constrained motion planning problem is defined as finding a continuous path
\[
\gamma : [0,1] \to \mathcal{M}, \quad
\gamma(0) \in Q_{\mathrm{start}}, \;\gamma(1) \in Q_{\mathrm{goal}},
\]
that moves the object from $x_{\mathrm{start}}$ to $x_{\mathrm{goal}}$ while satisfying the closed-chain constraints throughout.

\begin{figure*}[t!!]
    \centering
    \includegraphics[width=0.98\textwidth,trim={0 0 0 0},clip]{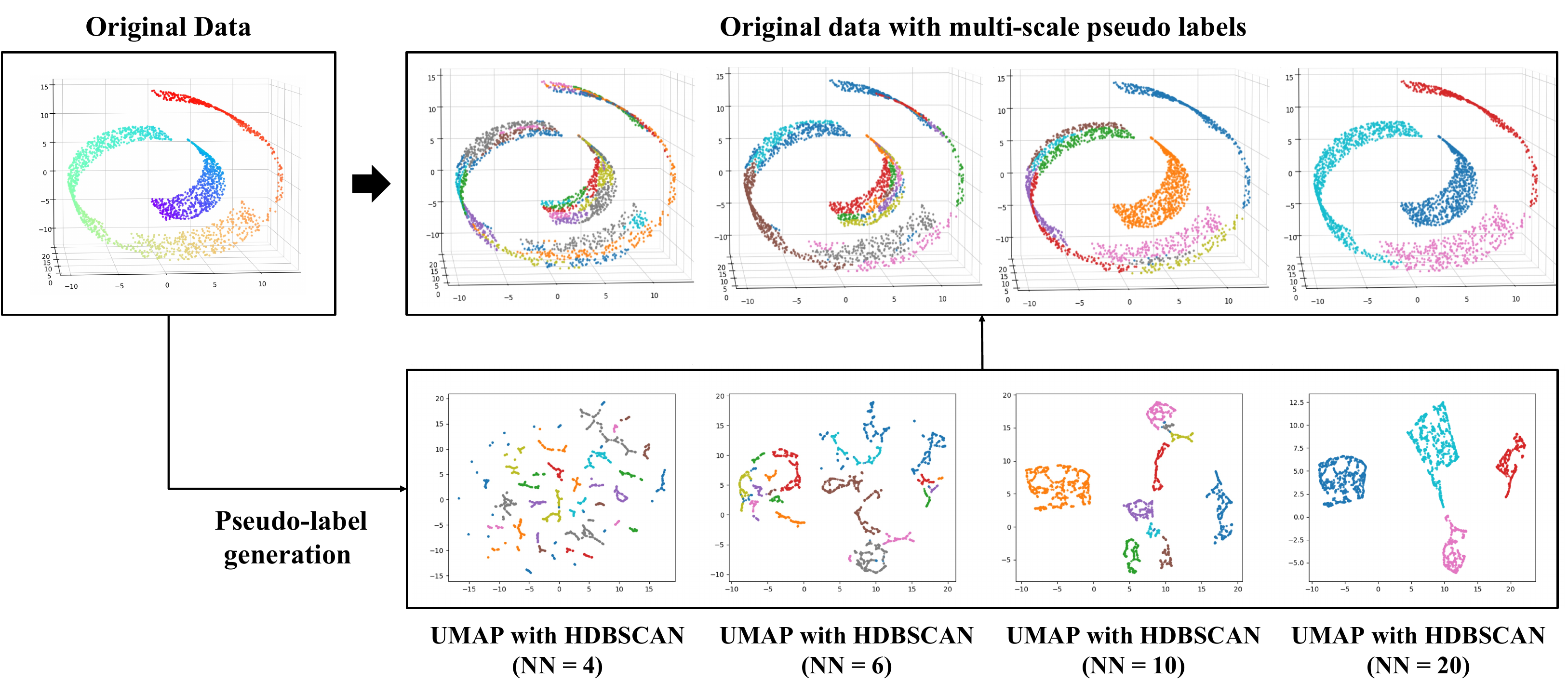}
    \caption{Multi-scale pseudo-label generation on a modified Swiss Roll dataset emulating constrained regions in robotic configuration spaces. (Top) pseudo-labels projected onto the original data. (Bottom) latent embeddings with clustering. Left to right: increasing nearest neighborhood (NN) scales from local to global connectivity.}
    \vspace{-0.2cm}
    \label{fig:swissroll}
\end{figure*}

\section{Learning Connectivity-Aware Representations for Constrained Motion Planning} \label{sec:method}

Robot configurations subject to task-specific constraints, such as closed-chain kinematics, form a low-dimensional manifold within the high-dimensional joint space. Feasible regions on this manifold are highly complex and often connected only through narrow passages, making direct analysis and planning difficult. While connectivity is formally defined as the existence of a feasible path between two configurations, in practice, it is more effective to reason about connectivity awareness, namely estimating which configurations are likely to be mutually reachable. Narrow passages, while mathematically continuous, are rarely sampled by planners and therefore behave as EMD components, highlighting the need for representations that capture connectivity structure and guide the choice of start–goal configurations.

To address this challenge, we construct a connectivity-aware representation of the configuration space (Fig.~\ref{fig:method}). The approach consists of two stages: (i) applying UMAP with multiple parameter settings and clustering the embeddings to generate multi-scale pseudo-labels of local connectivity, and (ii) training a neural network model with multi-scale contrastive learning using these pseudo-labels as supervision. The resulting latent space embeds mutually reachable configurations close together while separating disconnected ones, enabling efficient and reliable constrained motion planning.

\subsection{Generating Pseudo-Labels via Manifold Learning and Clustering}

To enable contrastive learning on robot configuration data, pseudo-labels are generated by embedding high-dimensional configurations into a latent space with manifold learning followed by clustering. Task-specific constraints restrict robot configurations to a low-dimensional manifold. This property motivates the use of manifold learning to capture the intrinsic low-dimensional structure of the feasible configuration manifold while preserving local connectivity.

Firstly, UMAP \cite{mcinnes2018umap} is applied to map the original joint configurations to a lower-dimensional latent space, preserving local neighborhood relationships and reflecting the connectivity of the constraint manifold. UMAP is chosen over alternatives such as VAE \cite{kingma2013auto} or t-SNE \cite{maaten2008visualizing} because it effectively captures both local and global manifold structure, handles disconnected regions robustly, and allows efficient mapping of new data points. To capture different scales of local structure, UMAP is applied with multiple parameter settings by varying the number of nearest neighbors and the minimum distance values. According to \cite{mcinnes2018umap}, larger neighbor sizes emphasize global structure while smaller ones highlight local neighborhoods, and the minimum distance parameter controls how tightly points are clustered in the latent space. Prior work has further shown that UMAP can improve clustering performance \cite{allaoui2020considerably}.

After obtaining latent embeddings at multiple scales, density-based clustering is performed using HDBSCAN \cite{mcinnes2017hdbscan}, an extension of the classical DBSCAN \cite{ester1996density} algorithm that automatically determines the number of clusters and identifies outliers. Unlike k-means\cite{cover1967nearest}, which assumes uniform cluster density and requires predefining the number of clusters k, HDBSCAN can naturally detect clusters of varying densities and separate disconnected regions. This is particularly important in constrained configuration spaces, where feasible regions can be disconnected or have narrow bottlenecks. HDBSCAN is applied with tuned parameters, adjusting minimum cluster size and samples to generate robust pseudo-labels that capture the structural connectivity of the configuration manifold.

Finally, clustering results from multiple latent embedding and clustering procedures are aggregated to generate multi-scale pseudo-labels, which serve as supervisory information for the next contrastive learning step. These pseudo-labels encapsulate structural information about the configuration space at various scales, enabling the latent representation to preserve connectivity and support effective constrained motion planning. The overall process, from manifold learning to clustering and pseudo-label aggregation, is illustrated in Fig.~\ref{fig:swissroll}. The modified Swiss Roll dataset emulates challenging structures of constrained configuration spaces, including narrow passages and disconnected regions. Multi-scale pseudo-labels are obtained by applying UMAP with clustering in the latent space (bottom) and then projected back onto the original data to highlight local connectivity (top).

\subsection{Contrastive Learning with Multi-scale Pseudo-Labels}

Contrastive learning aims to structure a feature-space by pulling together samples that are similar and pushing apart those that are dissimilar. Building on the pseudo-labels obtained in the previous stage a multi-scale contrastive learning strategy is proposed to exploit hierarchical connectivity information. At each scale, configurations assigned to the same cluster are treated as positives, while those from different clusters are treated as negatives. By jointly optimizing across scales, the learned representation preserves both global connectivity and local geometric structure of the feasible configuration space. This connectivity-aware representation provides a measure for selecting start and goal configurations likely to lie within the same connected region or favorable region, ultimately improving the feasibility and efficiency of constrained motion planning.

Formally, the framework trains a feature \(\phi: \mathbb{R}^n \rightarrow \mathbb{R}^d\), parameterized by neural network weights \(\theta\).  
Let \(\mathcal{X} = \{ x_1, x_2, \dots, x_N \}\) denote the dataset of robot configurations, and 
\(\mathcal{E} = \{ \varepsilon_1, \dots, \varepsilon_K \}\) the set of parameters used for multi-scale manifold learning and clustering.  
For each \(\varepsilon_k\), cluster assignments
\(\{ y_i^{(k)}\}_{i=1}^N\) are obtained for the data points \(x_i\).

At scale \(k\), the set of positive pairs for configuration \(x_i\) is defined as
\[
\mathcal{P}_i^{(k)} = \{ j \mid y_j^{(k)} = y_i^{(k)}, j \neq i \},
\]
while all remaining configurations are treated as negatives.  
Following SimCLR~\cite{chen2020simple}, null space augmentation~\cite{park2024constrained} is applied by generating perturbed samples $\tilde{x}_i$ that vary within the null space of the Jacobian while preserving task constraints. These augmented samples are always considered positive, promoting invariance to constraint-preserving variations.

The contrastive loss at scale \(k\) is defined using the InfoNCE objective, which is widely used in contrastive learning:
\begin{equation}
\mathcal{L}_{\mathrm{NCE}}^{(k)} = - \sum_{i \in \mathcal{I}} 
\log \frac{ 
\sum_{j \in \mathcal{P}_i^{(k)} \cup \{\tilde{x}_i\}} 
\exp\left( \frac{\mathrm{sim}(\phi(x_i), \phi(x_j))}{\tau} \right)}{
\sum_{j \in \mathcal{I} \setminus \{i\}} 
\exp\left( \frac{\mathrm{sim}(\phi(x_i), \phi(x_j))}{\tau} \right)},
\end{equation}
where \(\mathrm{sim}(u, v) = \frac{u^\top v}{\|u\|\|v\|}\) denotes cosine similarity and \(\tau\) is a temperature hyperparameter.  

The overall multi-scale contrastive loss is defined as a weighted sum across scales:
\begin{equation}
\mathcal{L}_{\mathrm{MLCL}} = \sum_{k=1}^K \lambda_k \mathcal{L}_{\mathrm{NCE}}^{(k)}, 
\quad \sum_{k=1}^K \lambda_k = 1, \quad \lambda_k \geq 0.
\end{equation}

After training, the learned embedding provides a continuous measure of connectivity between two configurations.  
Formally, given a mapping \(\phi: \mathcal{C}_{\text{free}} \to \mathbb{R}^d\), the representation is designed to capture connectivity-awareness:
\begin{enumerate}
    \item Configurations in connected and easily accessible regions are embedded close together.
    \item Configurations that are disconnected or lie in constrained bottlenecks are embedded farther apart.
\end{enumerate}

The connectivity measure between two configurations \(\mathbf{q}_s\) and \(\mathbf{q}_g\) is modeled as
\begin{equation}
P_\phi(\mathbf{q}_s, \mathbf{q}_g) = \exp\left(-\|\phi(\mathbf{q}_s)-\phi(\mathbf{q}_g)\|_2 \right).
\end{equation}

which generalizes beyond sampled configurations. This allows
connectivity-aware reasoning, particularly for selecting start and
goal IK solutions that are more likely to be connected under constrained motion planning tasks.

In the proposed method, the learned representation serves as a connectivity-aware measure prior to motion planning. Rather than randomly selecting inverse kinematics solutions, start–goal pairs are evaluated in the feature space, and candidates with smaller distances are prioritized. This pre-selection step avoids wasted search in EMD components, improves planning success rate, and reduces planning time by providing favorable initial conditions. 

\section{Experiments} \label{sec:experiments}

\subsection{Constrained Manipulation Planning Experiments}

The proposed algorithm is empirically evaluated on constrained manipulation planning tasks across multiple robotic platforms and diverse task constraints. Primary experiments are conducted on the Franka Panda arm, while additional evaluations on the Tocabi humanoid robot \cite{schwartz2022design} assess generality in higher-dimensional settings.

\subsubsection{Dataset Generation}

Directly sampling the constrained joint space is inefficient and misaligned with task-relevant regions. Instead, poses are sampled in the general task space to ensure comprehensive coverage of manipulation scenarios \cite{jang2022motion}. Dataset generation is performed in obstacle-free settings. For each sampled pose, inverse kinematics is solved using TRAC-IK \cite{beeson2015trac} to generate joint configurations, and only self-collision-free solutions are retained to construct the dataset. The associated null-space Jacobians are recorded for each valid configuration. This procedure is applied consistently across single-arm, dual-arm, and triple-arm tasks, and the resulting datasets are subsequently used for learning latent embeddings and null-space data augmentation.

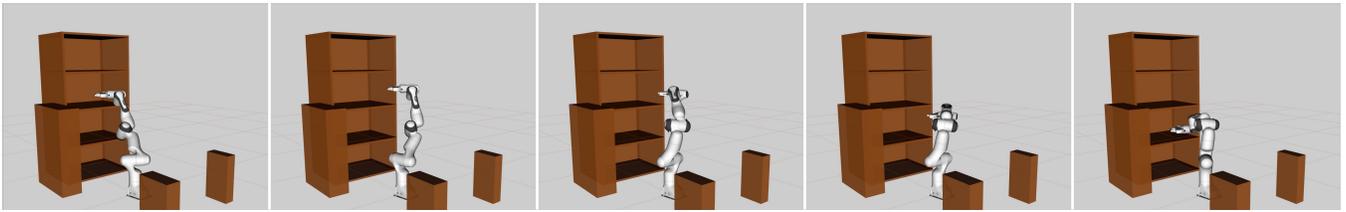
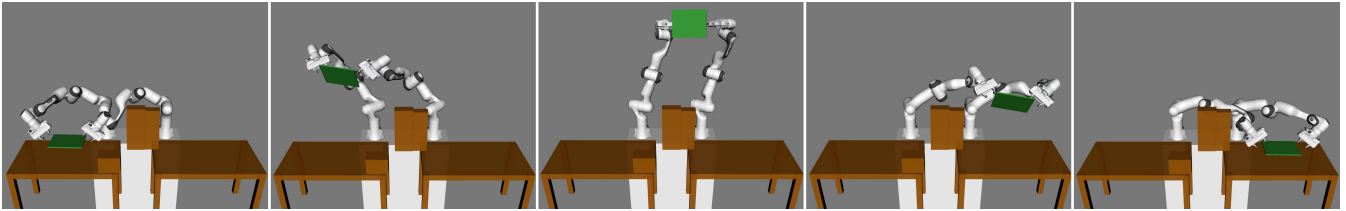
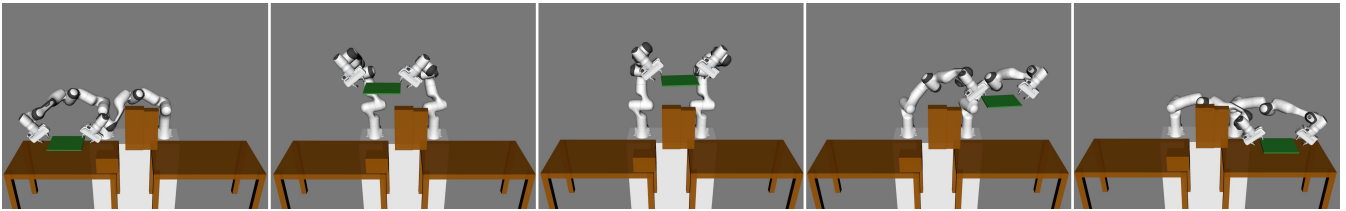
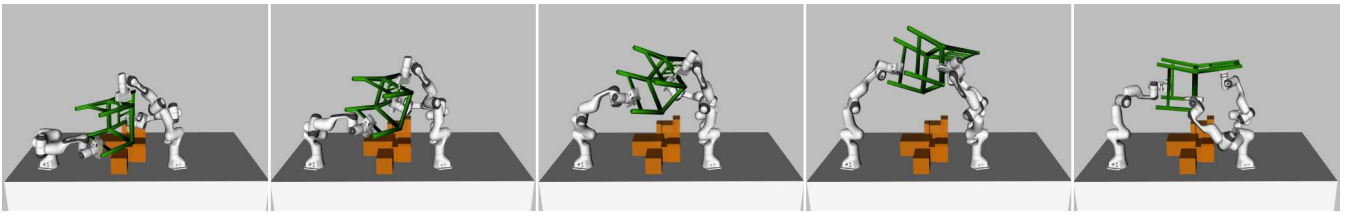
\begin{figure*}[t]
    \centering

    \begin{subfigure}{1.0\textwidth}
        \begin{tikzpicture}
            \node[anchor=south west,inner sep=0] (image) at (0,0)
                {\includegraphics[width=\linewidth,trim={0 6mm 0 0},clip]{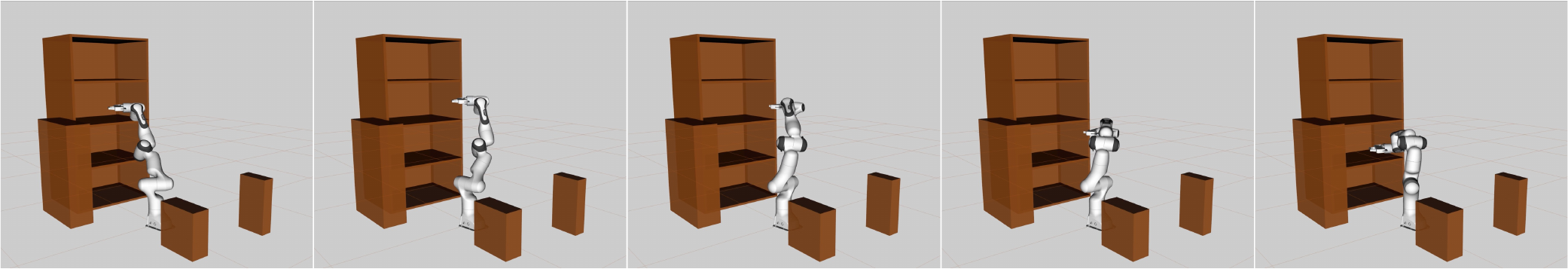}}; 
            \begin{scope}[x={(image.south east)},y={(image.north west)}]
                \foreach \x in {0.2,0.4,0.6,0.8} {
                    \draw[white, line width=1pt] (\x,0) -- (\x,1);
                }
            \end{scope}
        \end{tikzpicture}
        \caption{Single-arm manipulation: cup transfer task subject to fixed orientation constraints.}
    \end{subfigure}

    \vspace{0.1cm}

    \begin{subfigure}{1.0\textwidth}
        \begin{tikzpicture}
            \node[anchor=south west,inner sep=0] (image) at (0,0)
                {\includegraphics[width=\linewidth,trim={0 6mm 0 0},clip]{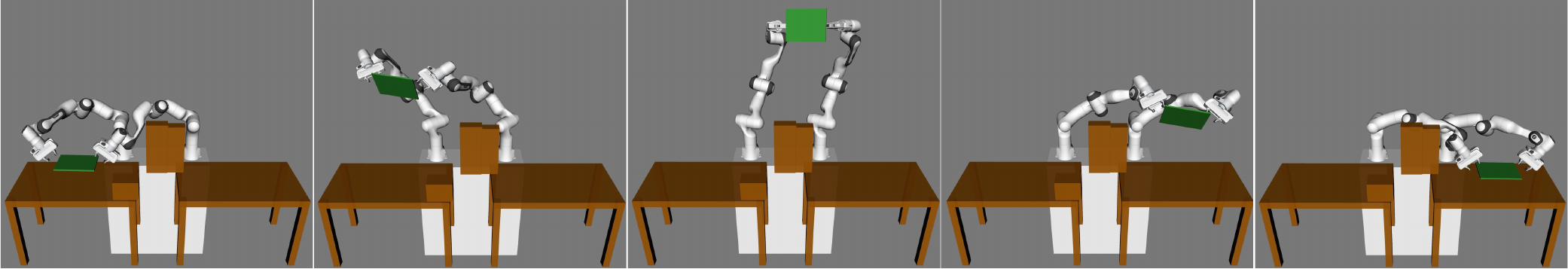}};
            \begin{scope}[x={(image.south east)},y={(image.north west)}]
                \foreach \x in {0.2,0.4,0.6,0.8} {
                    \draw[white, line width=1pt] (\x,0) -- (\x,1);
                }
            \end{scope}
        \end{tikzpicture}
        \caption{Dual-arm manipulation: tray transfer task performed under closed-chain constraints.}
    \end{subfigure}

    \vspace{0.1cm}

    \begin{subfigure}{1.0\textwidth}
        \begin{tikzpicture}
            \node[anchor=south west,inner sep=0] (image) at (0,0)
                {\includegraphics[width=\linewidth,trim={0 6mm 0 0},clip]{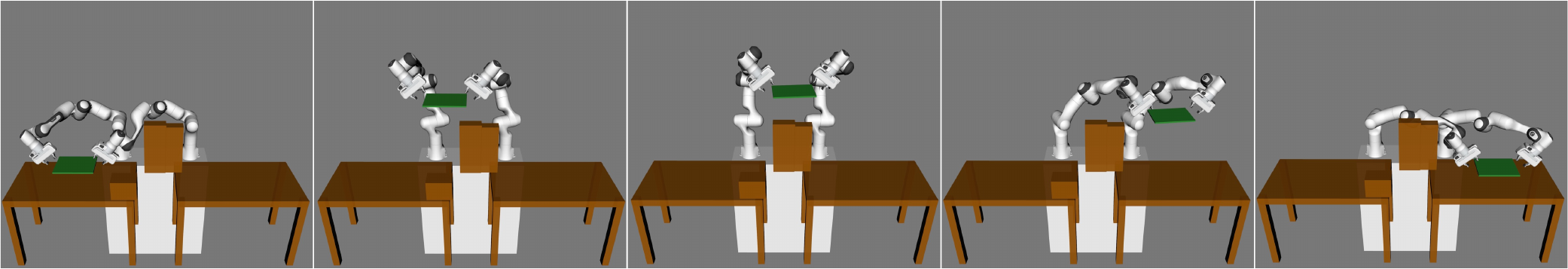}};
            \begin{scope}[x={(image.south east)},y={(image.north west)}]
                \foreach \x in {0.2,0.4,0.6,0.8} {
                    \draw[white, line width=1pt] (\x,0) -- (\x,1);
                }
            \end{scope}
        \end{tikzpicture}
        \caption{Dual-arm manipulation: tray transfer task performed under closed-chain constraints with fixed orientation.}
    \end{subfigure}

    \vspace{0.1cm}

    \begin{subfigure}{1.0\textwidth}
        \begin{tikzpicture}
            \node[anchor=south west,inner sep=0] (image) at (0,0)
                {\includegraphics[width=\linewidth,trim={0 6mm 0 0},clip]{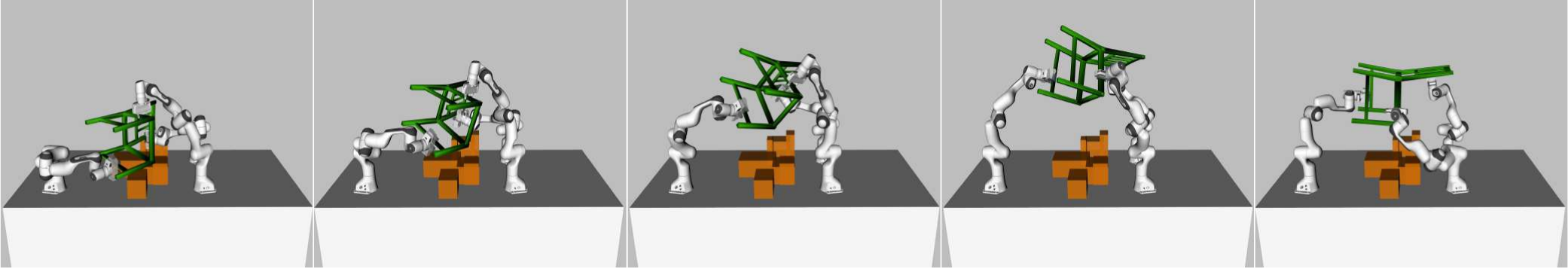}};
            \begin{scope}[x={(image.south east)},y={(image.north west)}]
                \foreach \x in {0.2,0.4,0.6,0.8} {
                    \draw[white, line width=1pt] (\x,0) -- (\x,1);
                }
            \end{scope}
        \end{tikzpicture}
        \caption{Triple-arm manipulation: chair rotation task under closed-chain constraints.}
    \end{subfigure}

    \caption{Experimental setup in simulation using Franka Panda robots across diverse manipulation tasks.}
    \label{fig:experiments_panda}
\end{figure*}

\begin{table*}[t]
    \centering
    \caption{Comparison of start–goal selection method on Franka Panda across diverse manipulation tasks (100 scenes, 100 second time limit). Metrics are the average planning time (successful trials only) and the success rates.
}
    \label{tab:planning_results_panda}
    % \resizebox{\textwidth}{!}{%
    \begin{tabular}{llcccc}
        \toprule
        & & \multicolumn{2}{c}{CBiRRT / 100 scenes / 100 sec} & \multicolumn{2}{c}{Latent Motion / 100 scenes / 100 sec} \\
        \cmidrule(lr){3-4} \cmidrule(lr){5-6}
        Task & Method & Planning time (sec) & Success rate (\%) & Planning time (sec) & Success rate (\%) \\
        \midrule
        \multirow{3}{*}{Fixed Orientation} 
            & Random         &  $6.599 \pm 11.533$  &  $92$ & $2.944 \pm 5.996$ & $94$  \\
            & Joint-Space    &  $6.986 \pm 8.744$   &  $\mathbf{100}$ & $8.230 \pm 16.891$ & $99$  \\
            & Feature-Space (Ours) &  $\mathbf{5.265 \pm 5.221}$  &  $\mathbf{100}$ & $\mathbf{1.808 \pm 3.702}$ & $\mathbf{100}$ \\
        \cmidrule(lr){1-6}
        \multirow{3}{*}{Dual Manipulation} 
            & Random         & $37.403 \pm 34.414$  &  $31$ & $4.700 \pm 7.904$  & $90$  \\
            & Joint-Space    & $47.211 \pm 26.268$  &  $41$ & $2.510 \pm 5.554$  & $\mathbf{100}$ \\
            & Feature-Space (Ours) & $\mathbf{30.865 \pm 28.612}$ &  $\mathbf{60}$ & $\mathbf{1.805 \pm 3.156}$  & $\mathbf{100}$ \\
        \cmidrule(lr){1-6}
        \multirow{3}{*}{\shortstack[l]{Dual Manipulation \\ with Fixed Orientation}}  
            & Random         & $33.988 \pm 31.297$  &  $27$ & $1.982 \pm 2.261$  & $46$  \\
            & Joint-Space    & $56.163 \pm 25.422$  &  $39$ & $2.092 \pm 2.510$  & $99$  \\
            & Feature-Space (Ours) & $\mathbf{32.552 \pm 29.684}$  &  $\mathbf{48}$ & $\mathbf{1.672 \pm 1.615}$  & $\mathbf{100}$ \\
        \cmidrule(lr){1-6}
        \multirow{3}{*}{Triple Manipulation} 
            & Random         & $75.350 \pm 0.000$   &  $1$  & $2.762 \pm 3.834$  & $30$  \\
            & Joint-Space    & $56.031 \pm 29.892$  &  $6$  & $1.906 \pm 2.533$  & $76$  \\
            & Feature-Space (Ours) & $\mathbf{32.370 \pm 23.106}$  & $\mathbf{10}$  & $\mathbf{1.500 \pm 1.371}$  & $\mathbf{84}$  \\
        \bottomrule
    \end{tabular}%
    % }
\end{table*}

\subsubsection{Experimental Setup}

The proposed method is evaluated on constrained manipulation tasks where end-effector poses are specified and feasible motions must be generated under kinematic and environmental constraints. A critical challenge in this setting is that multiple IK solutions may correspond to the same end-effector pose, yet not all start–goal pairs are mutually reachable. Consequently, selecting appropriate start and goal configurations is crucial for planning success. 

To comprehensively assess the approach, experiments are primarily conducted on the Franka Panda arm across four representative scenarios (Fig.~\ref{fig:experiments_panda}): a single-arm cup transfer, a dual-arm cooperative tray transfer, a dual-arm tray transfer with fixed orientation, and a triple-arm chair rotation. Each Panda arm has 7 degrees of freedom (DoF). In the dual-arm setting, closed-chain kinematic constraints reduce the dimensionality by 6 DoF, resulting in a 14–6=8 DoF constrained system. Adding a fixed orientation constraint further removes 2 DoF, leaving only 6 DoF. In the triple-arm setting, the combined 21 DoF are reduced by 12 through closed-chain constraints, yielding a 9 DoF constrained system. These reductions illustrate how constraint complexity increases planning difficulty across tasks.
For each scenario, planning is performed with both a traditional projection-based planner, CBiRRT \cite{berenson2009manipulation}, and a learning-based latent motion planner \cite{park2024constrained}, enabling assessment of whether the proposed method consistently improves performance across different planning frameworks. Each experiment is conducted on 100 randomized scenes, with three obstacles of varying size and position and a randomly sampled start–goal pose, under a planning time limit of 100 seconds. Note that these obstacles are not included during representation learning; the learned embedding is trained on obstacle-free configuration data and evaluated in environments containing previously unseen obstacles.

\begin{figure*}[t]
    \centering
    \begin{subfigure}{1.0\textwidth}
        \begin{tikzpicture}
            \node[anchor=south west,inner sep=0] (image) at (0,0)
                {\includegraphics[width=\linewidth,trim={0 10mm 0 0},clip]{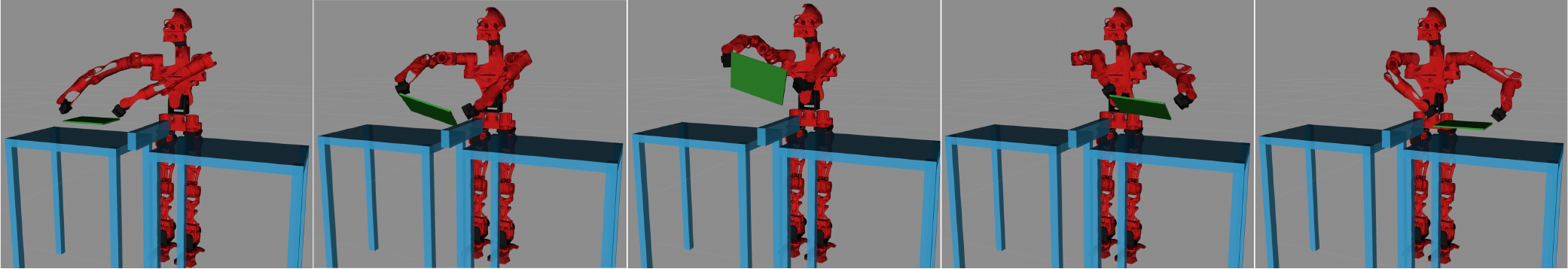}};
            \begin{scope}[x={(image.south east)},y={(image.north west)}]
                \foreach \x in {0.2,0.4,0.6,0.8} {
                    \draw[white, line width=1pt] (\x,0) -- (\x,1);
                }
            \end{scope}
        \end{tikzpicture}
        \caption{Dual-arm manipulation: tray transfer task performed under closed-chain constraints.}
    \end{subfigure}
    
    \vspace{0.1cm}
    \begin{subfigure}{1.0\textwidth}
        \begin{tikzpicture}
            \node[anchor=south west,inner sep=0] (image) at (0,0)
                {\includegraphics[width=\linewidth,trim={0 10mm 0 0},clip]{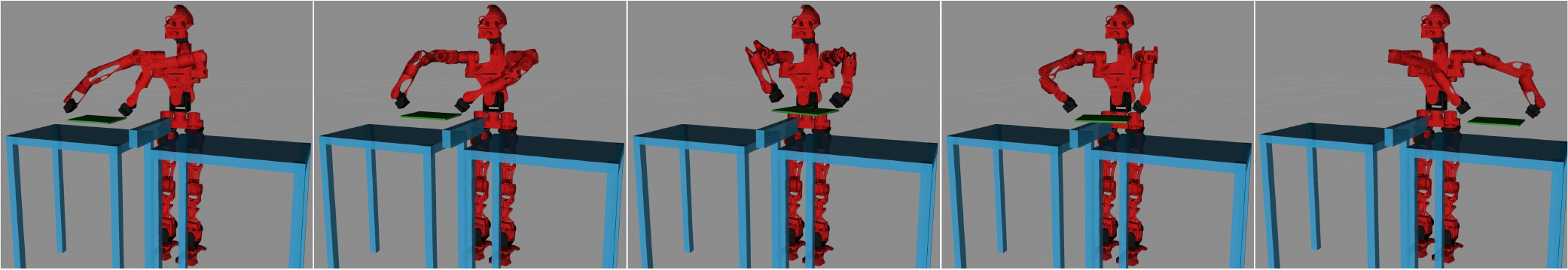}};
            \begin{scope}[x={(image.south east)},y={(image.north west)}]
                \foreach \x in {0.2,0.4,0.6,0.8} {
                    \draw[white, line width=1pt] (\x,0) -- (\x,1);
                }
            \end{scope}
        \end{tikzpicture}
        \caption{Dual-arm manipulation: tray transfer task performed under closed-chain constraints with fixed orientation.}
    \end{subfigure}

    \caption{Experimental setup in simulation using the Tocabi humanoid for dual-arm manipulation tasks.}
    \label{fig:experiments_tocabi}
\end{figure*}

Additional experiments on the Tocabi humanoid with 8-DoF arms are conducted to evaluate generality. In the dual-arm setting, 16 DoF are reduced to 10 by closed-chain constraints and further to 8 with fixed orientation constraints. As shown in Fig.~\ref{fig:experiments_tocabi}, two experiments on dual-arm manipulation and dual-arm manipulation with fixed orientation are conducted using CBiRRT under a 100-second time limit. This setup demonstrates that the proposed method scales to higher-dimensional arms and more complex kinematics while effectively handling redundancy.

Three strategies are compared for start–goal selection:
\begin{itemize}
    \item \textbf{Random:} Random selection of start and goal IK solutions.  
    \item \textbf{Joint-Space:} Selection of IK pairs minimizing Euclidean distance in joint space.  
    \item \textbf{Feature-Space (Proposed):} Selection of IK pairs minimizing distance in the learned feature space.  
\end{itemize}

\begin{table*}[t]
    \centering
    \caption{Comparison of start–goal selection method on the humanoid robot Tocabi using CBiRRT (100 scenes, 100 second time limit). Metrics are the average planning time (successful trials only) and the success rates.}
    \label{tab:planning_results_tocabi}
    \begin{tabular}{lcccc}
        \toprule
        & \multicolumn{2}{c}{Dual Manipulation} & \multicolumn{2}{c}{Dual Manipulation with Fixed Orientation} \\
        \cmidrule(lr){2-3} \cmidrule(lr){4-5}
        Method & Planning time (s) & Success rate (\%) &  Planning time (s) & Success rate (\%) \\
        \midrule
        Random         
            & $7.351 \pm 9.810$ & $34$ & $\mathbf{25.963 \pm 25.951}$ & $18$  \\
        Joint-Space    
            & $8.372 \pm 13.162$ & $\mathbf{98}$ & $35.778 \pm 25.378$ & $65$ \\
        Feature-Space (Ours) 
            & $\mathbf{6.909 \pm 8.584}$ & $\mathbf{98}$ & $38.341 \pm 26.326$ & $\mathbf{80}$ \\
        \bottomrule
    \end{tabular}
\end{table*}

\subsection{Results and Discussion}

Table~\ref{tab:planning_results_panda} summarizes the results of 100 constrained manipulation tasks using the Franka Panda robot with randomized obstacle placements and start–goal poses. Planning time is averaged only over successful trials. The method consistently improves both success rates and planning times across all tasks and environments. Compared to the joint-space distance baseline, the proposed method achieves up to $1.9\times$ higher success rates and reduces the average planning time to $0.43\times$. These gains arise from the connectivity-aware latent representation, which leverages multi-scale pseudo-labels and contrastive learning to avoid wasted computation in EMD components.
Importantly, these benefits are observed not only when integrated with the CBiRRT but also when combined with Latent Motion planner, demonstrating improvements for both data-driven and traditional planner.

On the Tocabi humanoid, Table II shows that planning is generally easier due to the higher redundancy of its 8-DoF arms. As noted in \cite{burdick1989inverse}, increasing the number of degrees of freedom tends to reduce the number of disconnected self-motion manifolds, which further improves configuration-space connectivity. In the dual-arm manipulation tasks, planning time decreases compared to both baselines, while success rates improve over the random baseline and remain comparable to the joint-space method. In contrast, in the dual-arm manipulation with fixed orientation tasks, planning time increases slightly, but success rates improve further, highlighting the benefit of connectivity-aware selection under stricter constraints. 

Although some failures remain, they appear to stem primarily from the inherent difficulty of exploring constrained manifolds within a limited time budget. Notably, success rates improve when a more capable planner is used with the same representation, suggesting that planner limitations, rather than representation deficiencies, contribute to the remaining failures. Nonetheless, learning accurate global connectivity remains challenging, particularly in high-dimensional configuration spaces with complex constraints.

\subsection{Implementation Details and Computational Efficiency}

The model consists of two main components: a four-layer MLP feature encoder with ReLU activations, and a two-layer MLP projection head with ReLU activations. Training is performed using the Adam optimizer with a learning rate of $10^{-4}$, batch size 1024, and 5000 epochs. For multi-scale pseudo-label generation, UMAP is applied with neighborhood sizes in the range ${3 \sim 50}$ and minimum distance values ${0.1, 0.2}$, followed by HDBSCAN clustering with minimum cluster size $20$ and minimum samples $10$. All experiments were run on a workstation with an Intel Xeon CPU and an NVIDIA RTX 4080 GPU.

In terms of efficiency, the preprocessing stage incurs a one-time cost of approximately 2 minutes for a dataset of 50,000 configurations in a dual-arm system with 14 DoF. Although the exact cost may vary depending on dataset size, robot configuration, and parameter settings, the overall preprocessing time remains modest. The inference time required to embed a configuration into the latent space is only 0.783 ms, which is negligible compared to the overall planning time. The model size also remains compact, at approximately 600 KB for the MLP model.

\section{Conclusion} \label{sec:conclusion}

This paper presented a connectivity-aware representation learning framework for constrained motion planning. By integrating manifold learning, clustering, and contrastive learning, the proposed method constructs embeddings that preserve both local and global connectivity, enabling improved start–goal configuration selection. Experiments on multi-arm manipulation tasks demonstrated consistently higher success rates and reduced planning times compared to baseline strategies. The current representation is learned in a task-specific manner and requires retraining when the underlying constraint manifold changes. Future work could focus on improving cross-task adaptability by conditioning on task or constraint parameters, learning shared structures across related manifolds, and developing hierarchical representations.

\bibliographystyle{IEEEtran}
\bibliography{references}
\end{document}